\documentclass[sigconf,screen]{acmart}

\makeatletter
\g@addto@macro\bfseries{\boldmath}
\makeatother

\usepackage{enumerate}
\usepackage{amsmath}
\usepackage{subfigure}

\AtBeginDocument{%
  \providecommand\BibTeX{{%
    \normalfont B\kern-0.5em{\scshape i\kern-0.25em b}\kern-0.8em\TeX}}}

\setcopyright{acmcopyright}
\copyrightyear{2020}
\acmYear{2020}
\acmDOI{10.1145/1122445.1122456}

\begin{document}

\title{Explaining Convolutional Neural Networks by Tagging Filters}

\author{Anna Nguyen}
\orcid{0000-0001-9004-2092}
\affiliation{%
  \institution{Karlsruhe Institute of Technology}
  \city{Karlsruhe}
  \country{Germany}
}
\email{anna.nguyen@kit.edu}

\author{Daniel Hagenmayer}
\affiliation{%
  \institution{Karlsruhe Institute of Technology}
  \city{Karlsruhe}
  \country{Germany}
}
\email{daniel.hagenmayer@student.kit.edu}

\author{Tobias Weller}
\affiliation{%
  \institution{University of Mannheim}
  \city{Mannheim}
  \country{Germany}
}
\email{tobi@informatik.uni-mannheim.de}

\author{Michael F{\"a}rber}
\orcid{0000-0001-5458-8645}
\affiliation{%
  \institution{Karlsruhe Institute of Technology}
  \city{Karlsruhe}
  \country{Germany}
}
\email{michael.faerber@kit.edu}

\renewcommand{\shortauthors}{Nguyen, et al.}

\begin{abstract}
Convolutional neural networks (CNNs) have achieved astonishing performance on various image classification tasks, but it is difficult for humans to understand how a classification comes about.
Recent literature proposes methods to explain the classification process to humans.
These focus mostly on visualizing feature maps and filter weights, which are not very intuitive for non-experts in analyzing a CNN classification.
In this paper, we propose \textsc{FilTag}, an approach to effectively explain CNNs even to non-experts.
The idea is that when images of a class frequently activate a convolutional filter, then that filter is tagged with that class.
These tags provide an explanation to a reference of a class-specific feature detected by the filter.
Based on the tagging, individual image classifications can then be intuitively explained in terms of the tags of the filters that the input image activates.
Finally, we show that the tags are helpful in analyzing classification errors caused by noisy input images and that the tags can be further processed by machines.
\end{abstract}

\settopmatter{printfolios=true} 
\maketitle

 \section{Introduction}

\begin{figure}[tb]
	\begin{center}	
		\includegraphics[width = 0.85\linewidth]{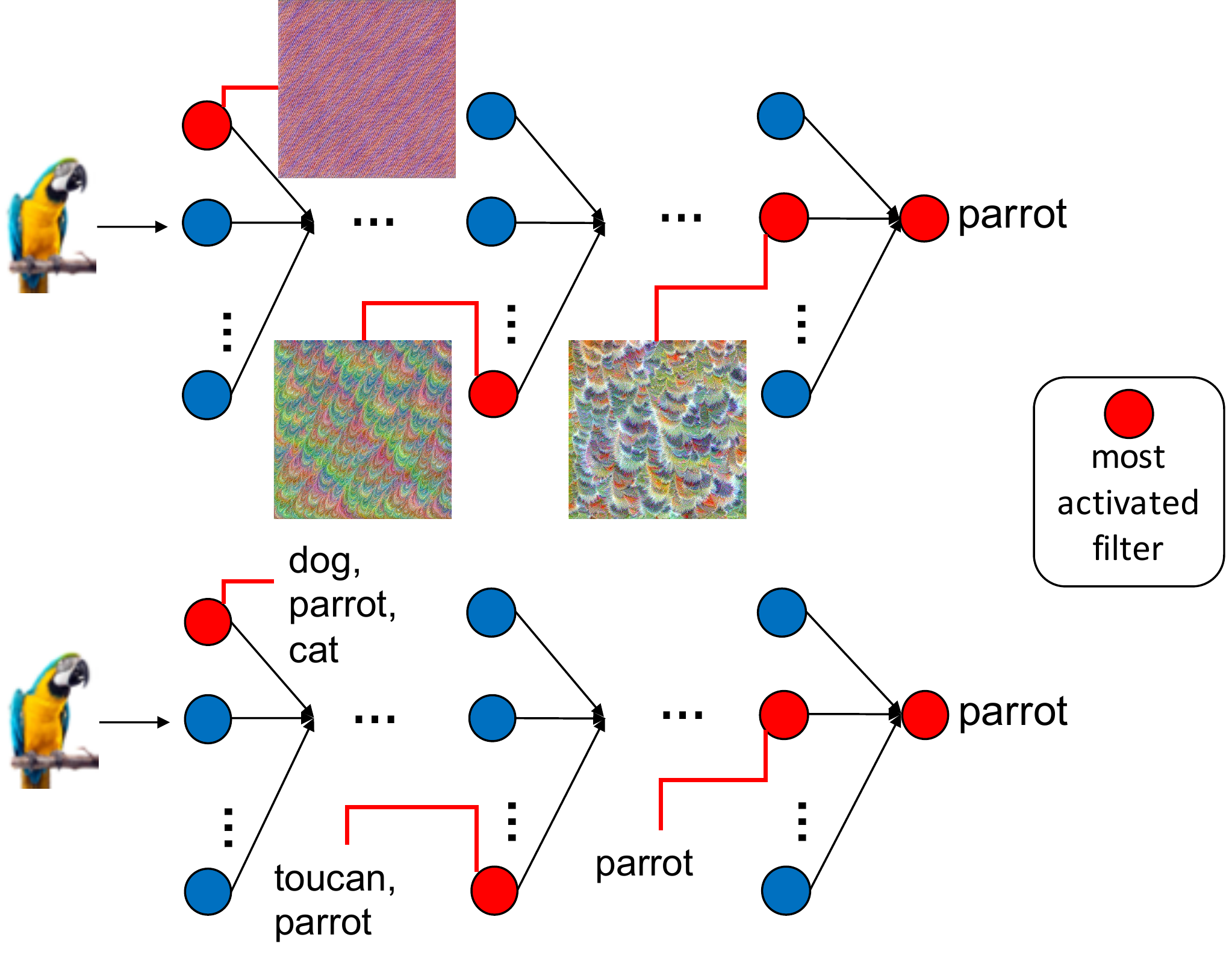}
		\caption{Explanations of Convolutional Filters. The upper part shows a visual explanation. The lower part contains an example of our tagging approach \textsc{FilTag}.}
		\label{explanation}
	\end{center}	
\end{figure}

\begin{figure}[tb]
	\begin{centering}
		\includegraphics[width = 0.8\linewidth]{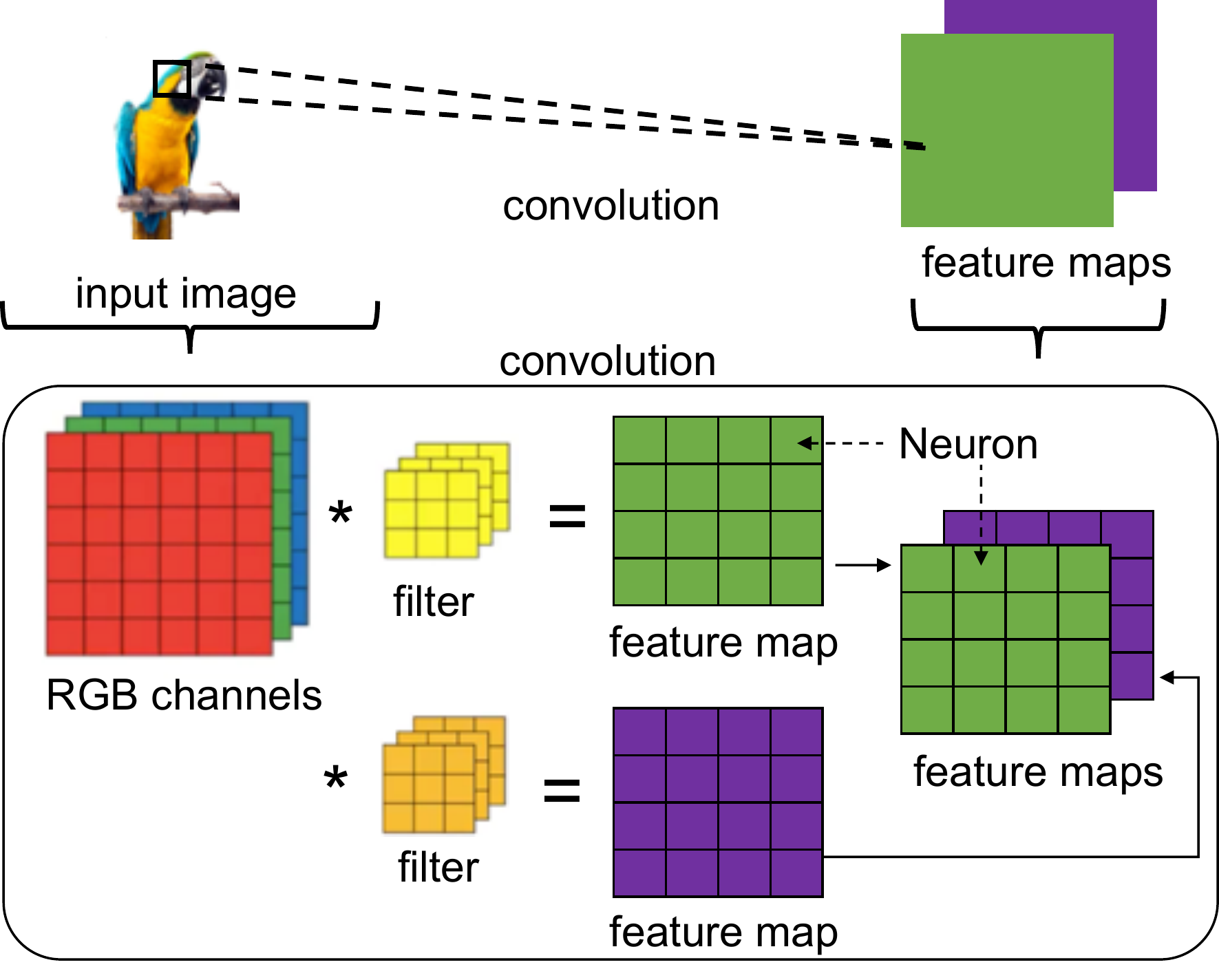}
		\caption{Terminology of a filter in a convolution.}
		\label{fig:definition}
	\end{centering}
		\vspace{-0.3cm}
\end{figure}

Deep convolutional neural networks (CNNs) are the state-of-the-art machine learning technique for image classification~\cite{VGG16}.
In contrast to traditional feed-forward neural networks, CNNs have layers that perform a convolutional step (see Figure~\ref{fig:definition} for the relations in a convolution).
Filters are used in a convolutional step which outputs a feature map in which activated neurons highlight certain patterns of the input image.
Although CNNs achieve high accuracy on many classification tasks, these models do not provide an explanation (i.e., decisive information) of the classifications. 
Thus, researchers recently focused on methods to explain how CNNs classify images.

\noindent\textbf{Related  Work.}
Some of the earliest works on explaining CNNs focus on visualizing the activations of individual neurons~\cite{olah2017feature}.
Still, these methods cannot explain more complex relationships between multiple neurons, as no human-understandable explanation is used. 
Therefore, Zeiler and Fergus~\cite{zeiler2013visualizing} visualize the filter weights to illustrate the patterns these filters detect.
However, these visualizations are based on the inputs of the layers to which the respective filter belongs to. Thus, only the filter patterns of the first layer can be directly associated with patterns on the input image of the network.
To overcome this, the method Net2Vec~\cite{FongV18} quantifies how concepts are encoded by filters by examining filter embeddings.
Alternatively, Network Dissection~\cite{BauZKO017} uses human-labeled visual concepts to bring semantics to the convolutional layers.
However, visualizations and embedding filters only explain the outcome of a model implicitly, whereas we assign explicit tags to filters which can be understood by non-experts. 
Most visualizations used for explaining CNNs are similar to the example in Figure~\ref{explanation}, which visualizes the most activated convolutional filters.
Clearly, such visualizations are difficult to understand on their own.
Adding an explicit explanation such as a semantic tag (e.g. dog, parrot, cat, or toucan) as shown in the bottom example would dramatically improve the explanation, including for non-experts.

\noindent
\textbf{Contribution.}
Our contribution is threefold.
First, we introduce \textsc{FilTag}, an automatic approach to explain the role of each convolutional filter of a CNN to non-expert humans.
We use the fact that each filter is dedicated to specific set of classes~\cite{zeiler2013visualizing}.
Indeed, the idea of \textsc{FilTag} is to quantify how much a filter is dedicated to a class, and then tag each convolutional filter with a set of particularly important classes.
The lower part of Figure~\ref{explanation} shows an example of what a CNN tagged in this way could look like.
In that example, the rightmost filter highlighted in red plays a role in classifying parrots, whereas the filter in the middle only plays a role in classifying birds in general, as both, toucans and parrots are both birds. This filter extract features that are specific to these classes (e.g. wings, feathers, etc.).
Second, our approach can also be used to explain the classification of an individual image.
In the example in Figure~\ref{explanation}, the classification of the input image as a parrot would be explained by the union of the tags of the activated filters, which are all animals, particularly tagged with parrot.
Third, \textsc{FilTag} is suitable to analyze classification errors.
We analyze our approach with thorough experimentation using ImageNet as a data set and using multiple CNNs, including VGG16, VGG19, and InceptionNet with pre-trained models. We focus in the experiment on VGG16 due to space limitations. All source code is available online\footnote{\url{https://bit.ly/3hjeMR4}\label{filtag}} and will be published on GitHub in case of acceptance.


\begin{figure*}[hbt!]
	\begin{center}	
		\includegraphics[width = 0.85\linewidth]{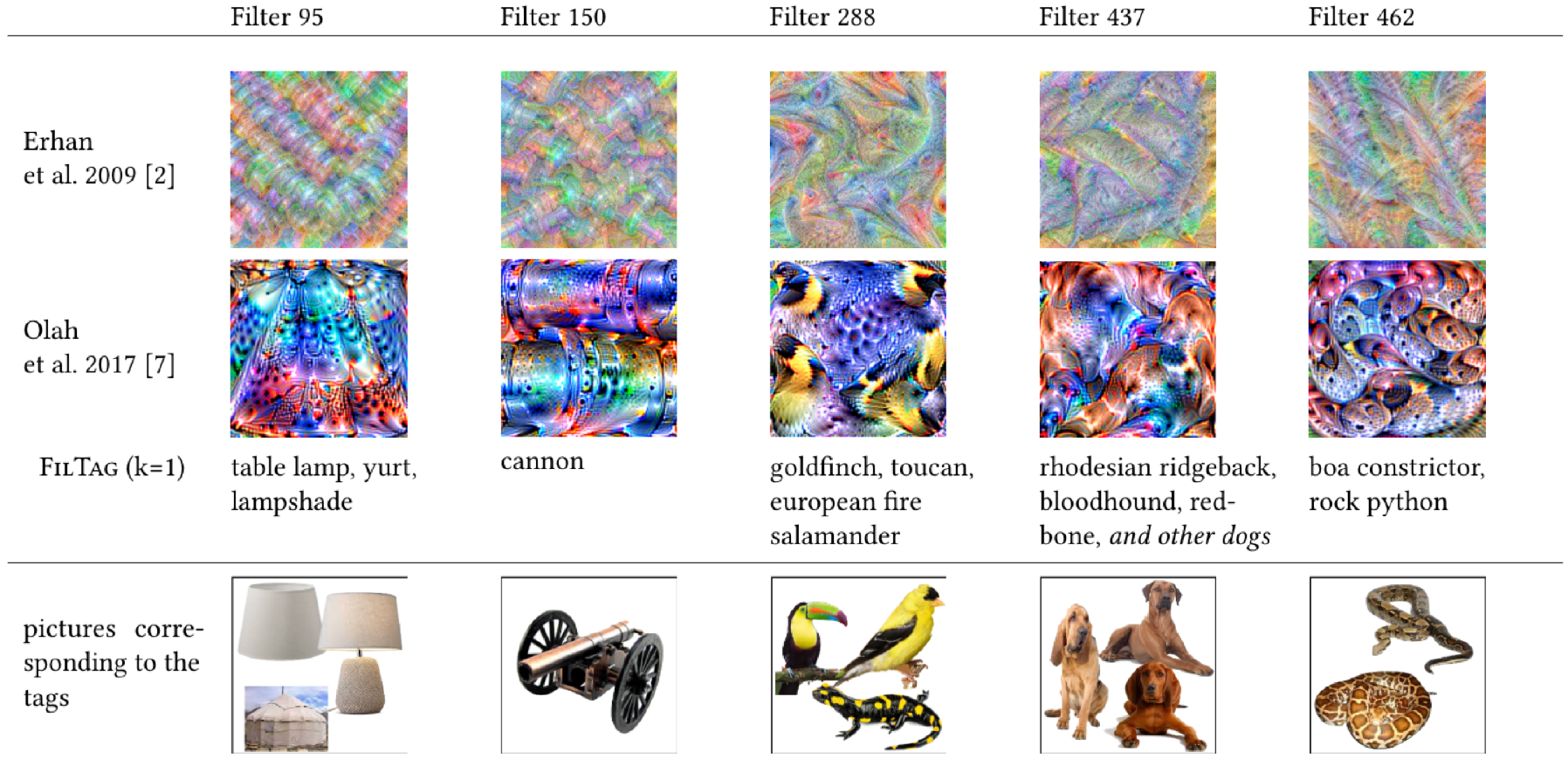}
		\caption{Comparison of filter explanations from the last convolutional layer from VGG16 \cite{VGG16}.} 
		\label{fig:experiment}
	\end{center}
	\vspace{-0.3cm}
\end{figure*}

\section{Approach}
\label{Approach}

In Section~\ref{computation}, we propose a method to provide explanations based on the role of each filter in a CNN (independent from concrete input images) using our concept of filter tags.
Then, in Section~\ref{metric}, we explain how a particular input image can be explained, namely in terms of the filters that it activates.

\subsection{Explanations of Filters}
\label{computation}

Our explanation of filters works in two steps.
In the first step, we quantify how much each filter is activated by images of each class.
In the second step, we use this information to tag the filters.

\noindent\textbf{Quantifying Filter Activations.}
Feature maps with high activations can be used as indication of the importance of the preceding filter for the input image~\cite{zeiler2013visualizing}.
Traditional explanation approaches focus on one image and therefore use the most activated feature map while our approach focuses on a set of images of the same class. \looseness=-1
\newline
Given a pre-trained CNN with a set of convolutional layers $M$ with its respective set of filters $I_{(\cdot)}$ and a labeled data set $D$ 
with labels $c\in C$ from a set of labels $C$, 
let $d\in D$ be an input image and $m\in M$ a convolutional layer.
First, we collect the activations in the feature map to get the importance of the filters regarding an input image, i.e. the output in the feature map for a given filter (see terminology in Figure~\ref{fig:definition}). 
Second, we scale these activations per layer between $[0,1]$. In scaling the activations, we ensure that no image is overrepresented with overall high activation values. We scale the activations per layer because each layer has its specific pattern compositionality of filters. For example, the first convolutional layers detect simple patterns such as lines and edges whereas the layers in the end detect compositional structures which match better to human-understandable objects \cite{zeiler2013visualizing}.
Let $a(m, i, d, j)$ be such a scaled activation in the $j$th element in the feature map calculated from image $d$ and filter $i\in I_m$ in convolutional layer $m$. In order to get a total activation value per feature map, we define $\bar a(m, i, d) = \frac{1}{n} \sum_j^n a(m, i, d, j), 0\leq \bar a(m, i, d)\leq 1,$ as the arithmetic mean of the scaled activations in a feature map 
where $n$ is the number of activations in the feature map.
We do this for all filters $i\in I_m$ and repeat these steps for all layers $m\in M$.
\newline
Next, we use the labels as the desired explanation. Let $d_c$ be an input image with label $c$. 
We define $z_c(m,i)= \frac{1}{{|D_c|}} \sum_{d_c}^{|D_c|} \bar a(m, i, d_c)$, $0\leq z_c(m,i)\leq 1$ as arithmetic mean of $\bar a(m, i, d_c)$ over one class $c$
where $|D_c|$ is the number of images in class $c$.
This way, $z_c(m,i)$ is the averaged value of all activations of the images in one class respective its filter $i$ in layer $m$.
Thus, we are able to rank the classes according to highest averaged activation of the filter per layer which will be the decision criterion for the labelling. We therefore compare the received values for each feature map.
We repeat these steps for all images in $D$ per label class. 

\noindent\textbf{Filter Tagging.}
We tag the filters according to their corresponding values received in $z_c(m,i)$ with the label of the input image class. We are interested in the feature maps with high activations of a certain class because they indicate important features associated with that class \cite{hohman2019}. We define two methods to select those feature maps per class and per layer (because of the mentioned complexity in different layers):
(i) $k$-best-method (choose the $k$ feature maps with highest activation values) and
(ii) $q$-quantile-method (choose the $q$-quantile of feature maps with highest activation values).
These tags serve as an explanation of what the filter does.
For example, in Figure~\ref{explanation}, the leftmost activated filter has the three tags \emph{dog}, \emph{parrot} and \emph{cat}, which suggests that this filter plays a role in recognizing animals.

\subsection{Explanations of Individual Classifications}
\label{metric}
While previous visual methods for explaining filters are difficult for humans to understand, textual assignment can lead to unambiguous explanations (as later seen in our experiments in Figure~\ref{fig:experiment}). To get an explanation given an input, we assume that the tags have a better information value with the classification of the CNN if the tags match with the classification output. 
Therefore, we want to measure the hit of the prediction with the tags in the most activated filters.
To do this, we determine the most frequently occurring labels for each image of a class according to the previous mentioned method using the metric Hits@$n$. Hits@$n$ measures how many positive label tags are ranked in the top-$n$ positions.
For example, in Figure~\ref{explanation}, the classification of the input image as a parrot is explained by its high activation of filters tagged with \emph{parrot}.


\section{Experiment}
\label{sec:4}

\subsection{Experimental Setup} 
\label{sec:4_1}

\textbf{Data Set.}
Following related work, we use ImageNet~\cite{ILSVRC15} data set from ILSVRC 2014 to conduct experiments on the introduced approach. This data set contains over one million images and $1,000$ possible class labels including animals, plants, and persons.
Each class contains approximately $1,200$ images. We use a holdout split, using $80\%$ of the images to tag the filters, while ensuring that there were at least $500$ images from each class in the set, and the remaining $20\%$ to test the explanations.

\noindent\textbf{Baseline.}
We compare our approach with two state-of-the-art visualization methods in explaining neural networks. 
The selection of the methods was based on their focus on feature visualization. One of the methods used provided the fundamental basis of visualization of features and uses minimal regularization~\cite{erhan}, the other method
uses optimization objectives~\cite{olah2017feature}.

\noindent\textbf{Implementation.}
We implemented our method in Python3 and used TensorFlow as deep learning library. The experiments were performed on a server with Intel(R) Xeon(R) Gold 6142 CPU@2.60 GHz, 16 physical cores, 188GB RAM and GeForce GTX 1080 Ti. We used pre-trained neural network models from Keras Applications. The filters of a VGG16 were explained in the experiments using the introduced method. VGG16 was used as CNN as it is frequently used in various computer vision applications. 
We also evaluated on VGG19 and InceptionNet but omit them due to page limitations. 


\subsection{Analysis of the Explanations}
\label{FilterLabelPres}
In this analysis, we want to study the explanations of the filters using $k$-best-method, with $k=1$, in order to provide a better comparison with the state-of-the-art methods since they frequently visualize the most activated feature map.
Figure~\ref{fig:experiment} shows exemplary the visual explanations of the baseline methods, and the tags of our approach \textsc{FilTag}.
As shown, the visual explanations of the baseline methods \cite{erhan,olah2017feature} do not provide satisfactory comprehension. At first sight, there is not much to understand. Considering our tags, one can imagine what the visualizations display. We additionally include pictures corresponding to our tags, to show the information value compared to only visualizations of the filters. Filter $95$ seems to recognize a lampshade especially a trapezoidal shape. Filter $150$ is only tagged with \emph{cannon}, i.e. the filter is specific for this class. 
Filter $288$ detects a head of a goldfinch especially with consideration of the yellow and black pattern. Filter $437$ and Filter $462$ recognize ears of brown dogs and the body of snakes, respectively. This information would be hard to retrieve without the tags. Even without considering the visualizations, one has a good impression of what a filter detects. For example, it is quite impressive that Filter $288$ detects this black yellow pattern which we can follow from the tags \emph{goldfinch, toucan}, and \emph{european fire salamander}. As well, Filter $95$ detects the trapezoid in \emph{table lamp, yurt}, and \emph{lampshade}.

In addition to comparing our method to the state-of-the-art methods in CNN explanations, we linked the tags to concepts from ConceptNet~\cite{SpeerCH17} to achieve a coarsening of common tags. ConceptNet is a semantic network with meanings of words. This comparison revealed that many tags have both visual and semantic commonalities (e.g., see Filter~437 in Figure~\ref{fig:experiment}, rhodesian ridgeback, bloodhound and redbone are all of type dog). 
Following this evaluation process, we manually reviewed 100 filters in the context of common visual and semantic commonalities. Here we found 88\% conformance with common tags in the filters.

\subsection{Impact of Hyperparameters}
\label{sec:4_2}

\begin{figure}
	\begin{center}	
		\includegraphics[width = 0.8\linewidth]{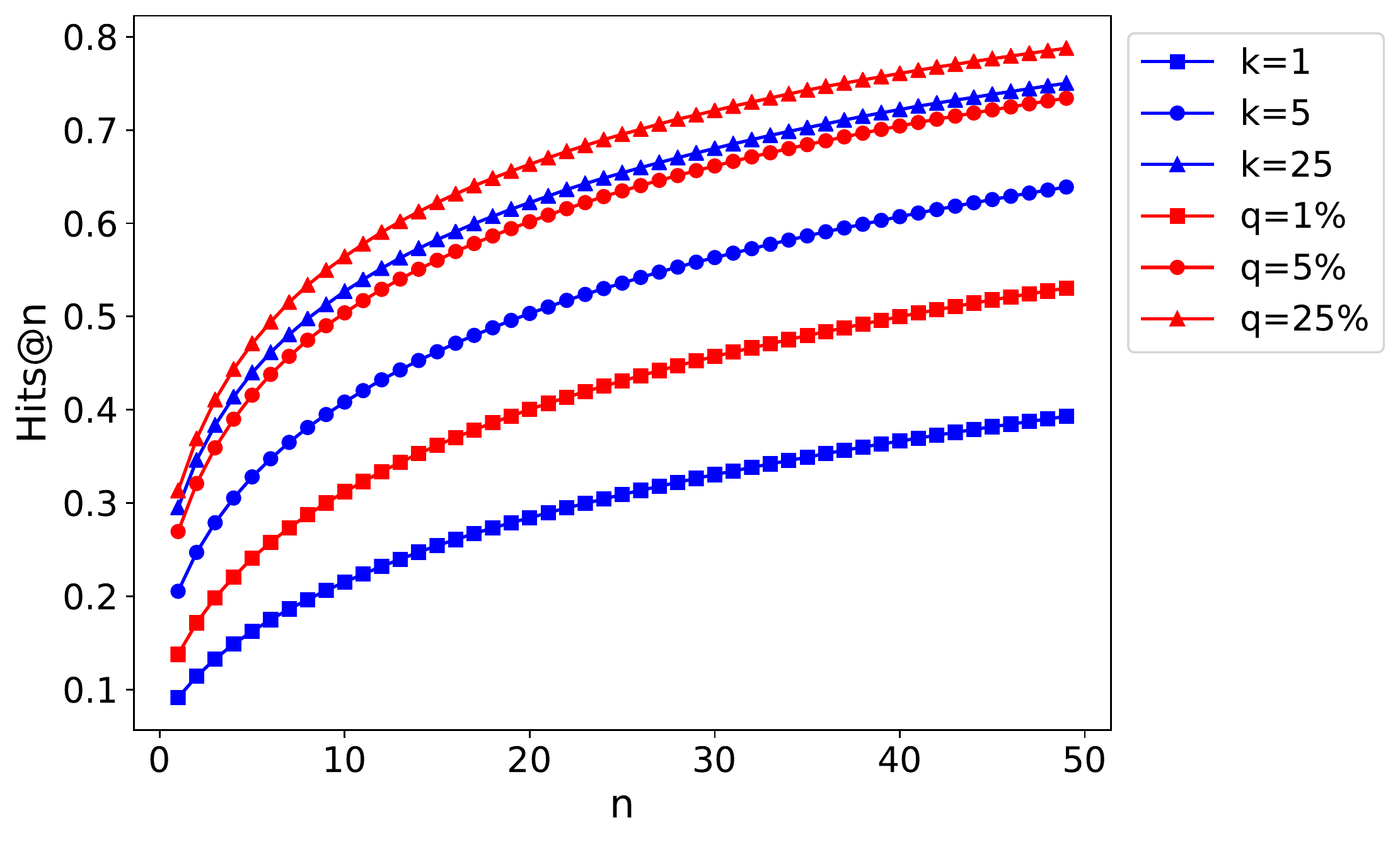}
		\caption{Hits@$n$ with different $k$ and $q$ on ImageNet}
		\label{hits}
	\end{center}
	\vspace{-0.3cm}
\end{figure}

In the following we evaluate which impact the hyperparameters $k$ and $q$ have on the correlation of Hits@$n$ and accuracy.
If the labels, and thus Hits@$n$, do not correlate with the output of the neural network, and thus with the accuracy, then the filters have not been tagged sensibly with our approach to gain an accurate explanation.
We will interpret Hits@$n$ and accuracy with different hyperparameters $k$ and $q$, respectively.
In Figure~\ref{hits}, we compute Hits@$n$ with the test set from ImageNet depending on $k$ and $q$. We can see that Hits@$n$ increases for increasing $k$, $q$ and $n$. For $q=25\%$ and $n=50$, we even get a hit rate of $80\%$ over all 1,000 object classes. 
This result shows that \textsc{FilTag} can be taken as a significant explanation for the classification. 
For example, we have observed that the class \emph{shoji} gets the highest hit rate of $98.47\%$ followed by the classes \emph{slot, odometer} and \emph{entertainment center} with also around $98\%$. 
This correlates with the likelihood of the best classes, which are exactly the same classes: \emph{shoji} ($81.22\%$), \emph{slot} ($92.30\%$), \emph{odometer} ($91.73\%$) and \emph{entertainment center} ($82.89\%$).
Likewise, Hits@$n$ also correlates with the accuracy of the worst classes, which are \emph{spatula}, \emph{schipperke}, \emph{reel}, \emph{bucket}, and \emph{hatchet}.
These results fit to the top-$1$ accuracy of VGG16 with $74,4\%$ for all classes.
The high correlation with Hits@$n$ and accuracy shows that the relevant features, labeled by our approach, are in fact detected from the images, which confirms the hypothesis that the tags are useful to generate explanations by means of our approach.
However, for larger values of $q$ we observed that the interpretability decreases because the number of tags increases for each filter. This makes it harder to find similarities between the classes. Thus, there is a trade-off between expressiveness for the classification and interpretability for the filters.\looseness=-1

\subsection{Using the Explanations}

\textsc{FilTag} can be used for error analysis using Hits@$n$. Taking misclassified input images, Hits@$n$ indicates if the most relevant filters were activated. If Hits@$n$ is high, we can assume that there are similar features of the misclassified class and original image. Analyzing the tags, we may find correlations in their semantics. 

\begin{figure}[tb]
    \subfigure[Mortarboard]{\includegraphics[height=0.3\linewidth]{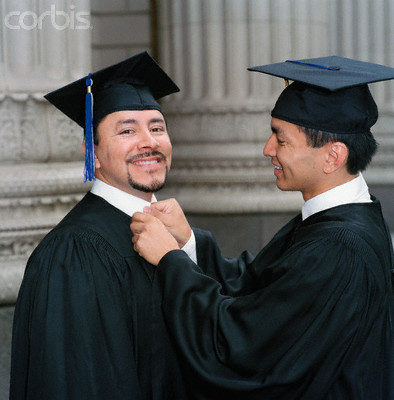}}
    \subfigure[Computer]{\includegraphics[height=0.3\linewidth]{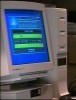}}
\caption{Example images from ImageNet}
\label{fig:image_example}
\vspace{-0.3cm}
\end{figure}

Figure~\ref{fig:image_example}~(a) shows an image of the class \emph{mortarboard} in ImageNet.
Using VGG16, the class \emph{academic gown} is predicted with a confidence of $83.8\%$, while the actual class \emph{mortarboard} is predicted with a confidence of only $16.2\%$.
Considering the image, we notice that both objects are part of this image, making this result reasonable. Reviewing the activated filters, we observe that filters tagged by \textsc{FilTag} with the tag \emph{mortarboard}, as well as with the tag \emph{academic gown}, are usually activated. 
As a result, we can verify that features are extracted from these two classes and used for prediction. This allows to give non-experts an understanding of the reason for the misclassification, as often features of the other class are extracted from this image.
Likewise, we can use the information to increase the number of images in which the mortarboard is the actual class but not in the main focus of the image, in order to continue learning the network to make the predictions more accurate.

Figure~\ref{fig:image_example}~(b) shows an image from the class \emph{computer}. This image is classified by VGG16 as \emph{cash machine} with a probability of $99\%$. 
Looking at the tagged filters, filters of the tags \emph{cash machine} are mostly activate, followed by \emph{screen, CD player}, and \emph{file}.
Considering Figure~\ref{fig:image_example}~(b) and having knowledge about the other images of the class \emph{computer} in ImageNet, the reason this image is not assigned to this class becomes clear. 
Generally, frontal images of a computer were used for the \emph{computer} class for learning. However, this image does not correspond to the same distribution. Thus, it is very difficult for the neural network to assign it correctly. Moreover, it is a very old computer, whereas the other images in ImageNet generally represent rather modern computers.
In order to classify this image correctly, further images showing old computers from the side have to be included to change the distribution and train the VGG16 to classify this image correctly.\looseness=-1

\section{Conclusion}
\label{sec:5}

We have introduced \textsc{FilTag}, an approach to provide human-under\-standable explanations of convolutional filters and individual image classifications. 
These tags can be used to query and identify specific filters that are relevant for feature detection. In contrast to state-of-the-art explanations, our approach allows for explicit, non-visual explanations which are more understandable for non-experts.
A limitation of our approach is the use of the class labels as tags to describe the filters. As a result, filters are not described in terms of specific objects such as ears, wings, or legs. 
We would like to address this limitation in the future by using ConceptNet and other knowledge bases to identify commonalities of the tags and thus add specific object descriptions to the filters. 
Furthermore, we plan to extend our evaluation for general classification tasks using CNNs.

\bibliographystyle{ACM-Reference-Format}
\bibliography{References}
\end{document}